\documentclass{article}

\usepackage{arxiv}

\usepackage[utf8]{inputenc} 
\usepackage[T1]{fontenc}    
\usepackage{hyperref}       
\usepackage{url}            
\usepackage{booktabs}       
\usepackage{amsfonts}       
\usepackage{nicefrac}       
\usepackage{microtype}      
\usepackage{graphicx}
\usepackage{soul}
\usepackage{bbold}
\soulregister\cite7
\soulregister\ref7
\soulregister\pageref7
\usepackage{algorithm,algorithmic}
\usepackage{amsmath}

\newcommand{\netname}{C$^3$Net}

\newcommand{\netnamespace}{\netname \hspace{1 mm}}

\begin{document}

\title{\netname: End-to-End Deep Learning for Efficient Real-Time Visual Active Camera Control
}


\author{
  Christos Kyrkou\thanks{ckyrkou@gmail.com,www.christoskyrkou.com}, \\
  \textit{KIOS Research and Innovation Center of Excellence}\\
  Department of Electrical and Computer Engineering\\
  University of Cyprus\\
  1 Panepistimiou Avenue, Nicosia Cyprus \\
  \{kyrkou.christos\}@ucy.ac.cy \\
}

\date{Received: date / Accepted: date}

\maketitle

\begin{abstract}
The need for automated real-time visual systems in applications such as smart camera surveillance, smart environments, and drones necessitates the improvement of methods for visual active monitoring and control. Traditionally, the active monitoring task has been handled through a pipeline of modules such as detection, filtering, and control. However, such methods are difficult to jointly optimize and tune their various parameters for real-time processing in resource constraint systems. In this paper a deep Convolutional Camera Controller Neural Network is proposed to go directly from visual information to camera movement in order to provide an efficient solution to the active vision problem. It is trained end-to-end without bounding box annotations to control a camera and follow multiple targets from raw pixel values. Evaluation through both a simulation framework and real experimental setup, indicate that the proposed solution is robust to varying conditions and able to achieve better monitoring performance than traditional approaches both in terms of number of targets monitored as well as in effective monitoring time. The advantage of the proposed approach is that it is computationally less demanding and can run at over $10$ FPS ($\sim4\times$ speedup) on an embedded smart camera providing a practical and affordable solution to real-time active monitoring.
\end{abstract}

\section{Introduction}
Real-time active vision systems (i.e., movable cameras with controllable parameters such as pan and tilt) can provide extended coverage, flexibility, and cost-efficiency compared to static vision systems running in the cloud. Active cameras can track targets (i.e. follow them) in order to record their movements and alert in case of an intrusion \cite{Micheloni2010VideoAnalysisPTZ}. Real-time operation on resource-constraint hardware is needed for such systems which are increasingly being used for various applications ranging from surveillance \cite{Angella2007OptDep}, mobile robots \cite{kyrkou:SCN:TCSVT:2018}, and intelligent interactive environments \cite{Wang:VisualComputer:2016}. 


Currently existing approaches for active vision decompose the problem into separate modules, namely detection, tracking, and control and employ different algorithms for the purpose of detecting targets and then following them \cite{Chen:NovelPTControl:2014,interaction:2019}. Such examples include motion detection, background modelling/subtraction, and lastly tracking by detection. The former two are widely used in static camera scenarios \cite{Hemangi:2015} due to their relative computational efficiency however, for active cameras that move and exhibit constant background change the latter methods are preferred since they are more widely applicable. Such techniques are often augmented with post-processing filters that increase the computational complexity making it difficult to use in embedded scenarios where the algorithm needs to run on the camera itself in real-time \cite{EmbeddedSmartCamerasBook2014}. Hence, existing methods that rely on hand-crafted features, motion models, and modeling of the camera views are not optimized for active vision scenarios e.g., tracking in the case of a camera mounted on a pan-tilt station \cite{Hemangi:2015}. Some approaches utilize a master-slave paradigm, but there has not yet been any attempt to deal with active monitoring of multiple targets in an end-to-end way. End-to-end learning approaches allow encapsulating all the intelligence into the machine learning algorithm thus can optimize all processing steps simultaneously and learning the best features to associate with camera control for visual active monitoring purposes.

\begin{figure*}[t]
	\centering
	\includegraphics[width=0.99\linewidth]{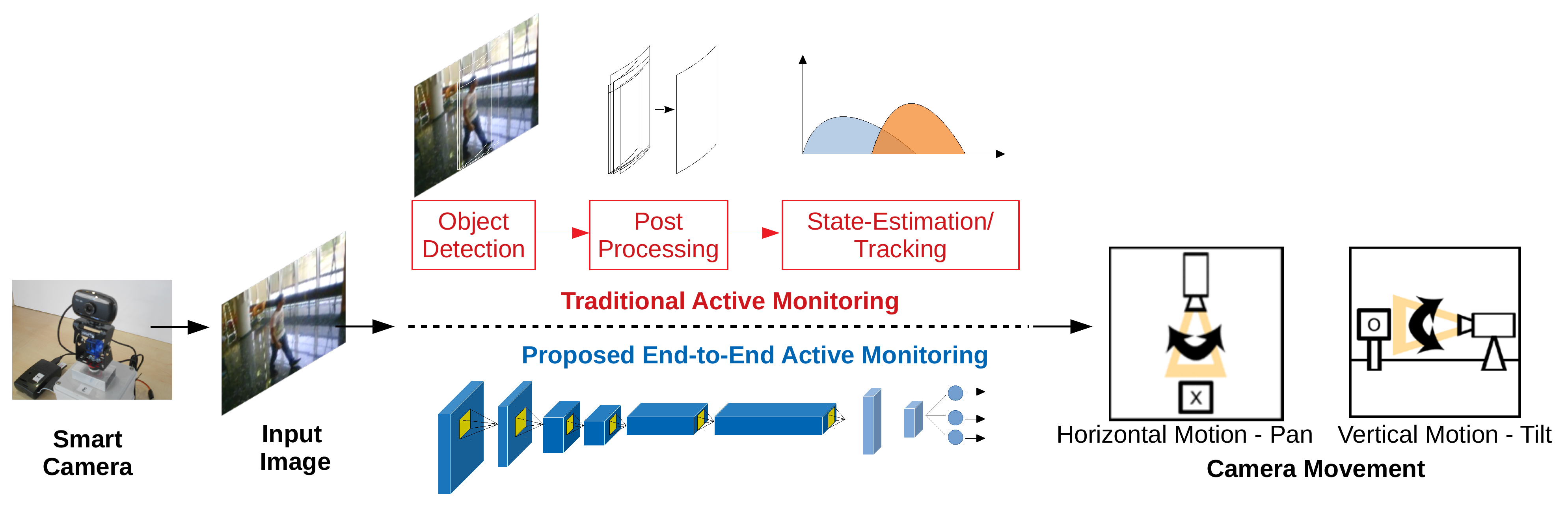}
	\caption{Traditional active monitoring Vs Proposed end-to-end deep visual active camera control.}
	\label{fig:endtoendoverview}
\end{figure*}

To deal with the aforementioned challenges, this paper investigates the appealing potential of end-to-end learning using Convolutional Neural Networks (CNN) \cite{PilotNet,VisualServoing2018,Miao_2018_CVPR} to develop an active vision system for monitoring people in surveillance and smart-environment applications. In particular the task of visual active monitoring is formulated as a supervised learning problem to be solved by deep learning in an end-to-end manner so that the low-level image features are directly associated with camera motion. In doing so it is possible to have a more compact system with reduced sub-modules to optimize and more efficient operation. An efficient \textit{C}onvolutional \textit{C}amera \textit{C}ontroller neural \textit{net}work, referred to as \netname, is designed and trained in an end-to-end manner to associate image features with control actions. To achieve the former an appropriate simulation framework and dataset for training and evaluating CNN-based controllers for active smart cameras is also introduced.

The effectiveness of the proposed \netnamespace has been verified by experiments through the simulation framework using image sequences from a publicly available PETS2009 \cite{PETS2009} dataset. Moreover, experiments were also performed under a real setup of a smart camera implementation and UAV acquired data. Results indicate that the proposed end-to-end approach is able to consistently monitor $1-3$ targets more on average than other methods and for a longer period of time. In addition, through experiments designed to test the robustness of the method it is validated that the network learns to focus on the majority of targets, without been given the bounding boxes and never explicitly trained to detect people. It also facilitates real-time performance for surveillance applications using resource-constraint platforms as it is computationally efficient as it has a small size and operates at over $10$ frames-per-second (FPS). 

The rest of this paper is structured as follows. Section II outlines relevant research and key areas of background material. Section III formally introduces the problem as well as the data collection and training procedures, and the proposed controller CNN. The experimental results and evaluation are presented in Section IV. Section V provides a discussion on possible extensions, while Section VI concludes the paper.

\section{Background and Related Work}
\subsection{Problem Overview}

The objective of active vision in contrast to a static camera setting is to change the control parameters of a vision sensor in order to maximize a visual-task-related performance objective such as following one or more targets that are located within the field-of-view (FoV) to improve the overall surveillance and monitoring capabilities. In particular this work deals with \textit{active monitoring} defined as the process of controlling the camera in order to position the center of its Field-of-View (FoV) at the center of mass of one or more targets in the image \cite{Bhanu:2011:DVSBook} (i.e., keep subject(s) as close to the center of the camera image as possible).  

Typically, detection algorithms \cite{ODreview:2019:TNNLS} that provide bounding box estimates centering around objects (Figure Fig. \ref{fig:endtoendoverview}) are used to guide the active vision systems. Bounding boxes used for localization and estimating target position can be affected by noise, occlusions, and even bad Non-maximum-suppression parameters \cite{ODreview:2019:TNNLS}. In addition, hand-crafted post-processing steps such as non-maximum suppression, used to remove superfluous detections, can lead to localization errors that affect the active monitoring performance \cite{Hosang_2017_CVPR,Bodla:SoftNMS:2017:ICCV}. Furthermore, tracking and state estimation approaches require fine tuning various parameters and can be negatively affected by the detection performance degradation. Therefore, there is a need to improve visual active monitoring methods for emerging applications that simultaneously require real-time performance, have battery limitations or operate in remote locations and with requirements for rapid deployment in temporary installations by providing computational cost that can lead to simpler active vision systems.

\subsection{Visual Active Monitoring}\label{activetrack}

Over the last years, deep neural networks, especially Convolutional Neural Networks (CNN), have improved the state-of-the-art in static object tracking/monitoring \cite{Feichtenhofer2017DetTrkDetTrk,Pflugfelder2017SiameseVisualTracking,REVAMP2019,interaction:2019}. Conventional solutions for active visual tracking tackle the problem by decomposing it into two or more sub-tasks \cite{Chowdhury:PTZcameras:TIP:2012}., i.e., object detection typically using a machine-learning-based classifier/detector, a tracking algorithms such as Kalman filter \cite{BewleySimpleTrack}, and a control output for the camera movement. Each task is optimized individually resulting in highly complex systems with many tuning parameters. Furthermore, this leads to difficulty in obtaining real-time performance on resource-constraint embedded camera systems. Different works have investigated the use of active cameras with one or more degrees of freedom for monitoring applications and are summarized next.

Initial approaches such as \cite{Lim2003ImageBasedPT}, followed a master-slave approach to track targets at a high resolution. One camera, the master, has a wide FoV and performs blob detection and uses a Kalman filter to track a target in an area and by projecting from image plane to world coordinates it controls (pan, tilt, and zoom) the other active camera, the slave, to follow a target. In contrast the goal of this work is to improve the tracking performance of a single camera agent so that it can autonomously follow all targets present in its FoV.

In \cite{Biswas2006DetTrackPTCam}, the authors proposed a camera control module to follow an intruder. A background model is used to detect motion in the scene while a Proportional-Integral (PI) controller moves the camera to follow the target. Illumination changes and crowded environments can affect the performance of such techniques. Moreover, the approach is only suitable for a single target.

A similar system was developed in \cite{Bernardin:FuzzyActiveCam:2007} where a combination of Haar cascades and color histograms are used to track face and upper body of a target in indoor environments. A fuzzy controller is used to steer the camera. The input to the fuzzy controller are the $x$ and $y$ position, as well as the size of the target object in the image. Likewise, the output of the fuzzy controller are the required pan, tilt, and zoom settings for the camera. The behavior of the system is determined by a set of rules connecting input values to expected outputs. The system can only track a single target and the hand-engineered rules can hinder its performance.

The system proposed in \cite{Dhillon.FaceActiveTrack.2009.CISIS} uses face detection to find a single face in the camera. It then extracts features for tracking and then feedback control loop is applied to vary the pan and tilt camera angles by certain steps based on the target velocity until it is centered. 

The approach in \cite{Dinh:RTactivePTZ:2009} uses complementary features and a multi-scale mechanism to follow the subject in front of the camera. In the first stage, a discriminative pre-filter automatically chooses the features which best separate object and background to build a confidence map which helps to remove all low confidence samples. In the second stage, a rough model of object appearance is built at the beginning and continuously updated according to all changes in pose, viewpoint, illumination. The accumulation over time can cause the model to fail with sudden changes in pose and cluttered background. Furthermore, this system again tracks only a single object.

In \cite{Haj:reacttivePTZ:2010} an Extended Kalman filter is used to jointly track the object position in the real world as well as estimating the intrinsic camera parameters. The filter outputs are used as inputs to two PID controllers (one for pan and one for tilt motion axis) which continuously track a moving target at a certain resolution. The focus of this work is on tracking only a single target however, and the objects are assumed to have a predetermined size. 

In \cite{Chen:NovelPTControl:2014} the authors propose a control approach for visual tracking to estimate and reduce the impact of different disturbances affecting the servoing system performance. The experiments where conducted with a pan-tilt camera following a single moving target. Furthermore, there is an assumption that the tracking module always performs ideally.

The active camera system proposed in \cite{Wang:VisualComputer:2016} is composed of multiple components in order to track a subject. The face of the target is identified through a face detection system and then a tracker is employed to estimate the targets motion across frames based on previous observations. An online learning approach is also used to learn the appearance model over time and reduce false detections. In addition, Gaussian Mixture Models are used to model the body movement in case the face detection fails. The final part is the controller of the pan-tilt camera which makes its decision of how to move based on where in the image the target is positioned and using a set of rules. This multi-component system can be difficult to tune and transfer to more resource-constraint systems that monitor multiple targets.

An end-to-end active tracker is proposed in \cite{Luo:endtoendRL:2017} for following a character in the VizDoom video game environment. It uses reinforcement learning to train a CNN with LSTM to output discrete movement actions. Even though they are somewhat realistic, these scenarios do not correspond to real-world use cases since evaluation is done in a video game environment whereas herein images of real environments from PETS2009 are used. Furthermore, the step-based output controls a video game player and is not suitable for the dynamic range need to control a pan-tilt camera. In addition, the control actions happen within the context and dynamics of a virtual world. In contrast, in this work the active monitoring problem is formulated in such a way as to be directly applicable to real-world camera systems.

In summary, it is evident from the literature that related works make excessive use of multiple modules composed of hand-crafted models and rules that must be tuned separately and in most cases track only a single target \cite{interaction:2019}. While there has been considerable progression in utilizing deep learning for static camera tracking there has been relatively few works dealing with deep learning for active smart camera systems. In addition, most of them do not consider important system requirements, such as real-time constraints on low-end hardware systems.

This work attempts to bridge the gap between the use of active cameras with deep learning algorithms by proposing an end-to-end learning approach to simultaneously built an implicit detector and controller for cameras with pan and tilt motion capabilities. In contrast to existing works e.g. \cite{Luo:endtoendRL:2017,interaction:2019} that follow a single target our goal is to give the same priority to all targets and attempt to follow as many as possible without a specific focus a single target. In addition, it does not require to be given an anchor target to follow and thus can be used in generic scenarios where the goal is to monitor targets in an area. Overall, the approach eliminates the need to rely on bounding box predictions which can be noisy and face difficulties with dense targets.

\section{Deep Active Visual Monitoring}
\subsection{Approach Overview}\label{sec:approachoverview}

The problem of end-to-end active visual control of a camera through a single image, as shown in Fig. \ref{fig:endtoendoverview}, is formulated as a regression problem. The input is an image $I$ from the camera sensor with resolution $I_x\times I_y$, and the output is a motion control vector $\vec{M}$. The control vector corresponds to the pan and tilt motion that the camera will have to perform in order to position the target(s) close to the center of the image. A Convolutional Camera Controller neural network (\netname) is trained end-to-end to learn a control function $f$ such that $\vec{M} = f(I)$.

\begin{figure}[t]
	\centering
	\includegraphics[width=0.79\linewidth]{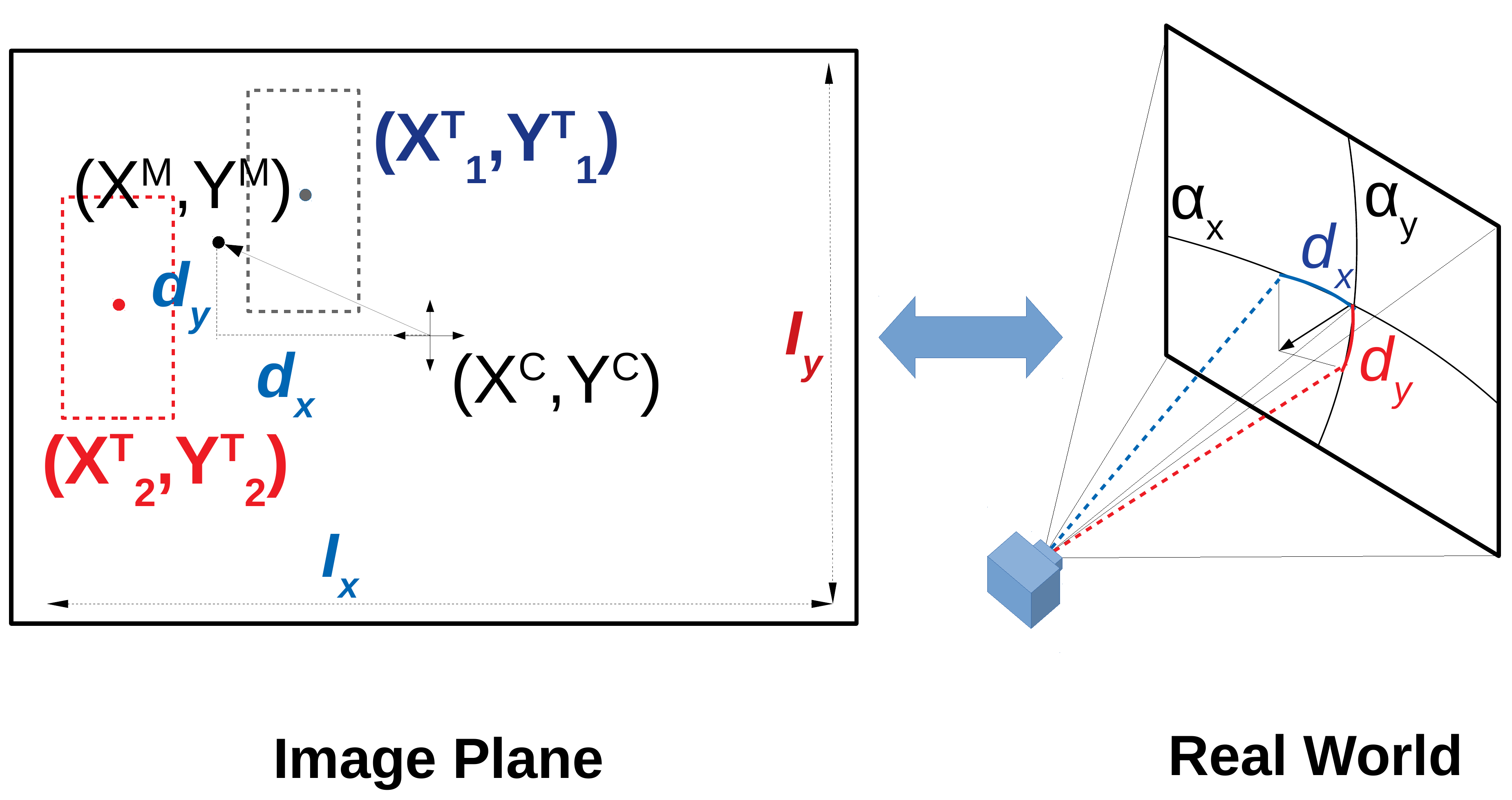}
	\caption{Mapping from Image Plane to Real-World: (Left) Displacement between image center and target center of mass. (Right) Camera angle of views associated with motion at each axis.}
	\label{fig:model}
\end{figure}

A pinhole camera model and rectilinear lens \cite{Haj:Beyond_Static:2011} are assumed, as illustrated in Fig. \ref{fig:model}. Under this assumption the pixel distances of the image center to the camera center are analogous to the angle that the servo motors have to move in order to position the target at the center of FoV \cite{Chen:NovelPTControl:2014,Dhillon.FaceActiveTrack.2009.CISIS,kyrkou:SCN:TCSVT:2018,Salih:2012:CameraGeometry}. In addition, the motion of the camera when following targets in its FoV is bounded by the half of its horizontal and vertical angles of view ($\alpha_x,\alpha_y$). Accordingly, the motion of the camera in the the pan and tilt axes (denoted by $M_x$ and $M_y$) can be calculated with respect to the viewing angles \cite{Salih:2012:CameraGeometry}. This can be done by first calculating the distance of the camera center from the target(s) center of mass ($X^M,Y^M$) for the horizontal and vertical direction ($d_x,d_y$), and then use Eq. \ref{eq:Motion_Est} to associate the pixel distance to the camera angles. Hence, during the learning process the objective is to map the input image to normalized pixel displacement values ($d_x/I_x,d_y/I_y$) which will then be used to calculate the corresponding angle displacements in the pan and tilt axes. Predicting offsets instead of angles effectively bounds the output making it easier for the network to learn, and decouples the learning process from camera specific parameters.

\begin{align}
M_x \rightarrow \dfrac{d_x}{I_x} \times \dfrac{\alpha_x}{2}, M_y \rightarrow \dfrac{d_y}{I_y} \times \dfrac{\alpha_y}{2}
\label{eq:Motion_Est}
\end{align}

\subsection{Convolutional Camera Controller Network (\netname)}

\begin{figure*}[t]
	\centering
	\includegraphics[width=0.99\linewidth]{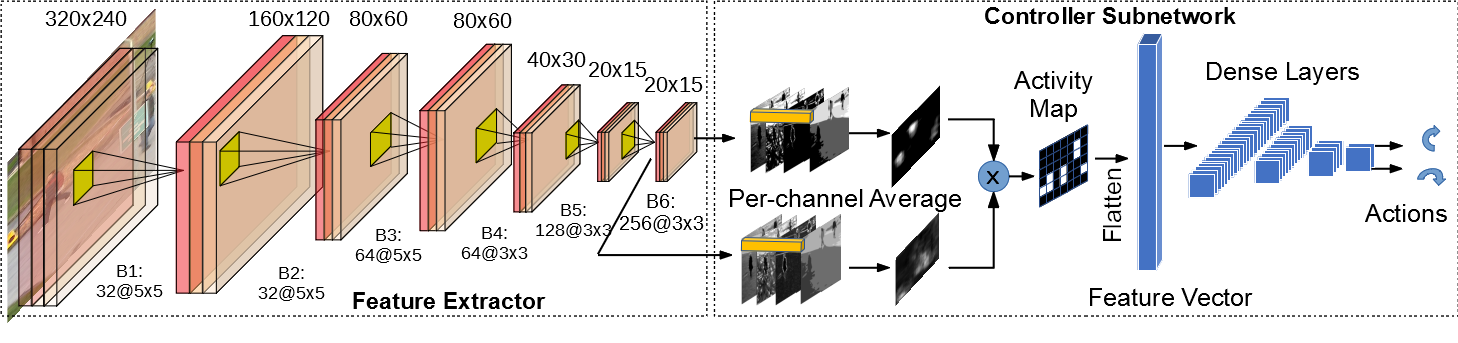}
	\caption{\netnamespace Architecture: Conceptually comprised of a feature extractor and a controller that condenses the high level semantic information to infer an action.}
	\label{fig:CNN}
\end{figure*}

\netnamespace is a suitable CNN trained to perform the regression task and estimate the camera motion displacement. Figure \ref{fig:CNN} shows the network architecture which is comprised of two main parts, a feature extractor made up of convolutional layers and a motion controller that summarizes the feature maps in order to calculate the final output values. Since the system is trained end-to-end it is difficult to distinguish the parts of the network that function primarily as feature extractor and which serve as controller. The input to the CNN is an RGB image which has normalized pixel values between $[0\dots1]$. The network can process images of $320\times240$, but any other image size can be provide after it is resized.

\textbf{Convolutional Feature Extractor:} The layers were designed to perform feature extraction and were chosen empirically through a series of experiments that varied layer configurations. There are $5$ major blocks each comprised of a convolutional layer with \textit{Relu} activation and batch normalization layer. Furthermore, dropout is applied at the middle of the sub-network to combat overfitting with a rate of $0.2$. Overall, the feature extractor is designed to be inherently computationally efficient to support use in embedded smart cameras for local control and decision making. To reduce the computational cost the image is down-sampled early on and a relatively small number of filters is used in the convolutional layers. Note that this is not an explicit object detector that produces bounding boxes, but rather implicitly learns the features to focus on through the controller supervision.

\textbf{Controller Subnetwork:} The controller subnetwork is responsible to map the extracted image features to control actions. It is comprised of layers with predetermined functionality as well as fully-connected layers. The custom layers serve to condense the information from the feature extractor essentially encoding our prior knowledge on the problem. The custom layers perform a per channel average of the feature maps extracted from blocks 5 and 6 (Fig. \ref{fig:CNN}). The reason of using the last two layers is that they encode more semantically meaningful information thus can provide better indication for the presence of objects in the image. The response of each feature map is averaged over the channels and then multiplied element-wise emphasizing the overlapping regions. This activity map is then reshaped into a vector and given as input to the fully connected layers. The idea behind this is to leverage multi-level features to first encode an activity map with the presence of objects over an image grid and then provide a small vector to the fully connected layers to reduce parameter count. There are $4$ fully connected layers with $100$,$50$,$10$ and $2$ neurons respectively to map the feature vector to motion controls. There is a dropout in between the dense layers and all layers have a \textit{leakyrelu} activation function. The output of the controller subnetwork is further processed through a hyperbolic tangent activation that bounds the output between $[-1,\dots,1]$. Overall, \netnamespace has a total of $\sim500,000$ parameters which requires $\sim4$MB, resulting in a lightweight network that can run even on low-end CPUs.

\subsection{Active Camera Data Generation}\label{sec:simframe}

\begin{figure}[t]
	\centering
	\includegraphics[width=0.99\linewidth]{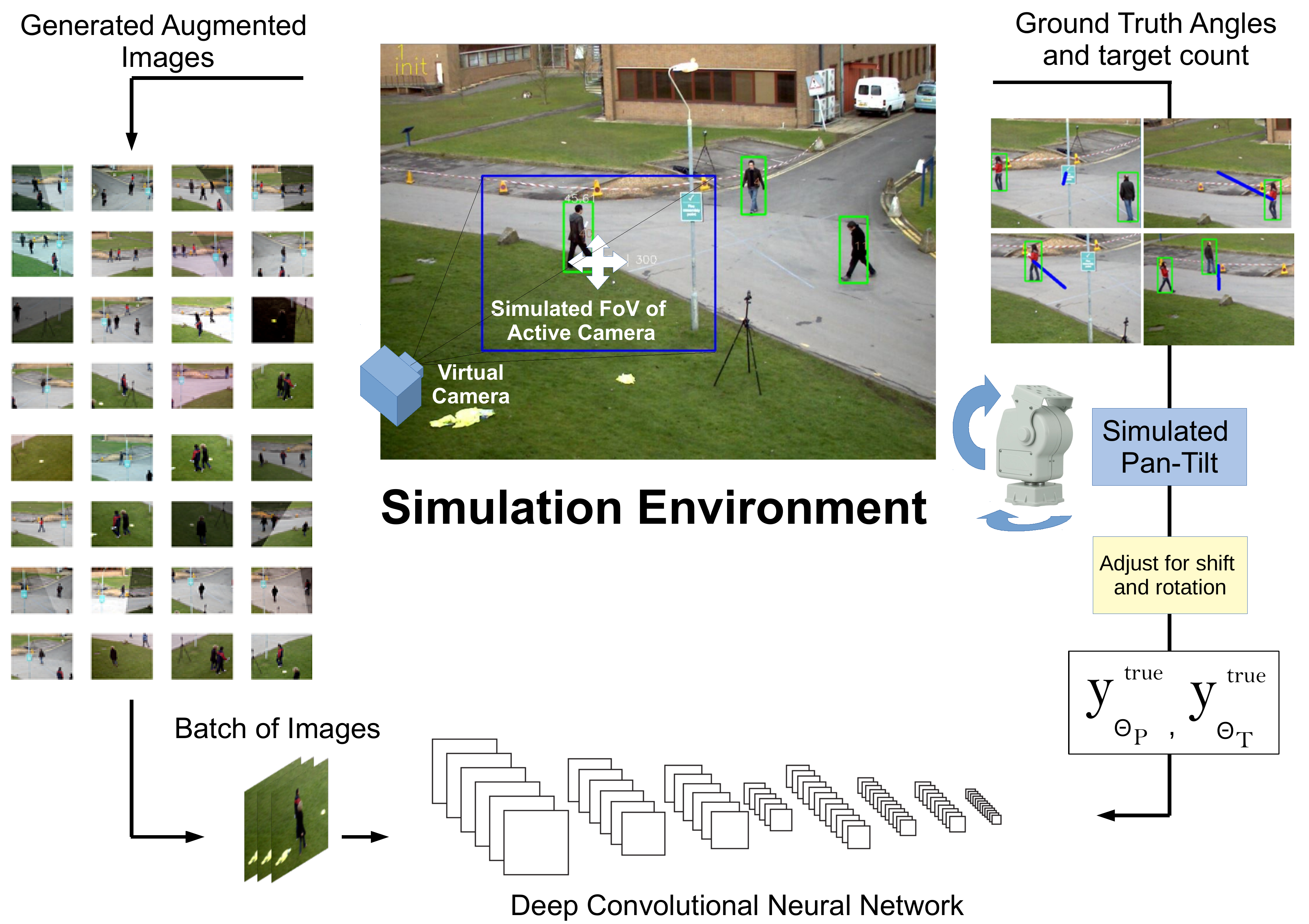}
	\caption{Overall framework for simulating virtual active camera movements and tracking, and extracting ground truth data.}
	\label{fig:framework}
\end{figure}

Proper training and testing data are necessary to train a deep CNN regressor for the visual active monitoring task. To the best of the authors knowledge there is no publicly available dataset for active vision applications with ground truth camera controls. For this reason an existing multi-person image dataset, such as the PETS2009\cite{PETS2009}, traditionally used for static tracking \cite{BoBo:2015:RMT:2789116.2789125} is re-purposed to develop a simulation framework. The particular dataset is selected over more recent ones as it has multiple targets, larger image frames where targets only occupy a small region, and has ground truth bounding box annotations. 

Overall, the framework allows for i) simulate the behaviour of active cameras using real-world images, ii) Capturing and storing multiple frame sequences with ground truth data that can be used for bounding box, density, and camera control, iii) evaluate performance of active vision algorithms in a realistic environment in similar conditions and controlled experiments. The simulation framework shown in Fig. \ref{fig:framework} simulates the movement of a virtual active camera with pan-tilt control with a fixed FoV by moving it within a larger image frame. By cropping a fixed region from the original frame, it effectively restricts the active camera to a limited part of the overall image frame. These crops take $\sim15\%$ of the original image. Obtaining the ground truth data for each image crop is a straight forward procedure if bounding box annotations are provided with the dataset, as is the case with PETS2009. First, the center of mass is calculated for all objects within each crop. The difference between the centre of the mass of the annotated people and the centre of the cropped region, in $x,y$ directions, dictates the two control signals for that frame as per Eq. \ref{eq:Motion_Est}. Alternatively if no bounding boxes are provided it is possible to use a trained object detector to provide candidate boxes and train in a weakly supervised manner. The ground truth of images without objects is set to $(0,0)$, meaning no movement. It is worth noting the same exact values used to move the camera in the simulated environment can be used to move the camera in the real-world when multiplied with the camera horizontal and vertical viewing angles. Thus the framework provides a valuable complementary data source for developing and benchmarking data-driven methodologies for smart camera applications.

\begin{figure}[t]
	\centering
	\includegraphics[width=0.99\linewidth]{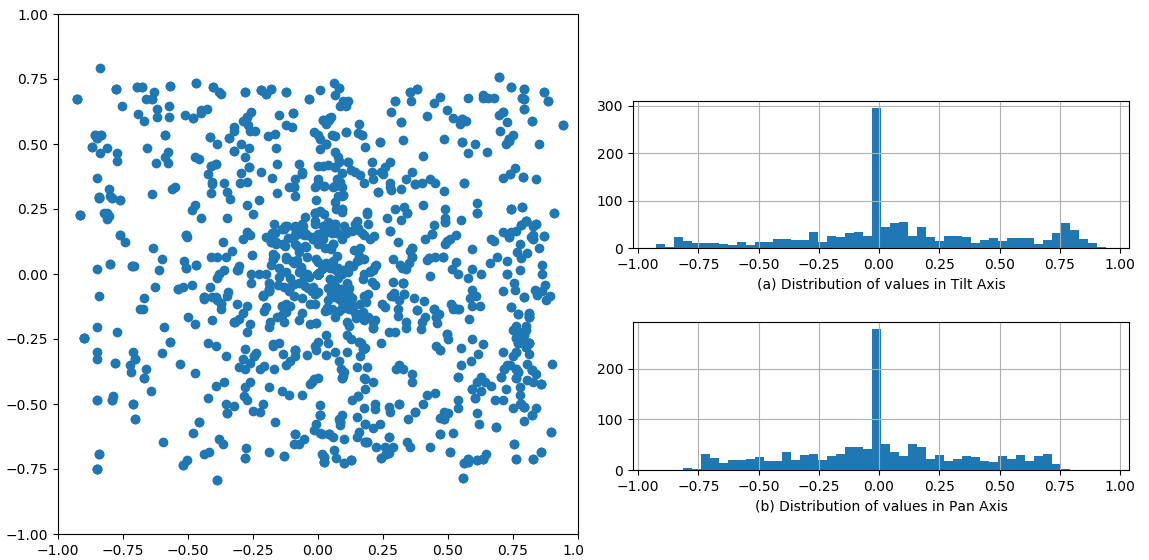}
	\caption{Distribution of the pan and tilt values. The majority of values are close to zero so selective sampling is used to balance the dataset.}
	\label{fig:angledistribution}
\end{figure}

A single sequence from the PETS-2009 \cite{PETS2009} database, with ground truth bounding boxes, was selected as the basis to generate training data. A dataset of over $4500$ frames was generated using the simulation software in order to train \netname. Starting at various positions in random frames in the sequence the changes in displacement of pan and tilt axes are calculated according to Eq. \ref{eq:Motion_Est} so that the targets will be positioned at the camera image center in the next frame. Hence, the training set effectively contains single images sampled from the sequence, paired with the corresponding ground truth future pan and tilt displacement. The pan and tilt displacement values are normalized between $-1$ and $1$ with regards to the maximum motion of the camera in both axes. For this work images of $320\times240$ are considered, which have been used in similar studies \cite{Bernardin:FuzzyActiveCam:2007}, but it is also trivial to extract different resolution images. 

An important part of data preparation is balancing the data. During the data collection process depending on the motion of successive frames, the target displacement values may not change significantly in both axes. As a result, the distribution of camera motion values might be imbalanced as shown in Fig. \ref{fig:angledistribution}. To ensure that the learning algorithm does not overfit by producing small values around zero, during training the data in each batch is sampled to contain high as well as low values.

The whole data is randomly split into $60\%$ for training the CNN regression model, and $40\%$ of the data was used for validation purposes. In addition, augmentations are probabilistically applied to the images to increase the variability and combat over-fitting. The augmentation strategy included some transformations on the image pixels such as blurring and sharpening, color-shifting, illumination changes. Geometric transformations such as translations and horizontal flip were also performed with appropriate adjustment of the target pan and tilt values. The combinations of all these augmentations resulted in a variety of novel images that were used for training.


\subsection{Network Training}
The objective of the learning process is to regress a motion vector $\vec{M}$ corresponding to motion in the pan and tilt axes. Accordingly the euclidean distance loss function in Eq. \ref{eq:loss} is employed for learning the camera controls between ground truth $y$ and predictions $\hat{y}$. The \textit{Keras} deep learning framework \cite{keras} with \textit{Tensorflow} \cite{tensorflow} running as the backend is used for the training of the CNN regressor. The network was trained using a GeForce Titan Xp, on a PC with an Intel $i7-7700K$ processor, and $32$GB of RAM. The Adam optimization method was used for training with a learning rate step-decay approach starting from an initial learning rate of $0.001$, and decreasing it by $0.5$ when encountering a plateau. The CNN regressor is trained for $300$ epochs with a batch size of $128$ and $50$ batches per epoch resulting in $6400$ augmented images per epoch.


{\normalsize  
\begin{align}
L=\dfrac{1}{N_B}\sum^{N_B}_{j=1} \bigg[(y^{j}_{d_{x}}-\hat{y}^{j}_{d_{x}})^2 +(y^{j}_{d_{y}}-\hat{y}^{j}_{d_{y}})^2 \bigg]^{1/2}&&
\label{eq:loss}
\end{align}
}%

\subsection{Weighted Moving Average Filtering}
A form of momentum is introduced to smooth out the outputs of the neural network and avoid being misled by a single frame, that is based on a weighted moving average (Eq. \ref{eq:rolavg}). It is grounded on the assumption that the target motion will not change rapidly between frames. By incorporating a simple form of memory into the system the previous actions are also considered which reduces the impact of possible outliers. A window of $3$ most recent outputs is maintained ($K=3$) which was empirically found to represent a good trade-off between having a smoother control output but not being influenced too much by passed control outputs. The weight $w$ can be determined through different policies. Herein, the weight factor is determined as shown in Eq. \ref{eq:rolavg}, where $t$ equals $K$ is the most recent output that is weighted the highest. Each frame the network produces output $\hat{y}$, and the moving average produces the filtered output $\overline{y}$.

\begin{align}
\overline{y} = \sum_{t=1}^{K} \hat{y}_{t}\times w_{t}, \text{ where } w_{t} = \dfrac{t}{\sum_{z=1}^{K}z}
\label{eq:rolavg}
\end{align}

\section{Evaluation}
To compare the proposed method a traditional active tracking pipeline is implemented as shown in Fig. \ref{fig:endtoendoverview}. It follows the tracking-by-detection approach used by related works for active camera control \cite{Dina2017CNNTrackDet,BewleySimpleTrack}. In this paradigm an object detector is used to localize the targets and a tracker is applied to filter the detections over time. Finally based on the filter target positions the camera is moved to position the targets to their center of mass. We compare the different approaches with respect to their monitoring efficiency with metrics outlined in section \label{sec:perf_metrics}. Also through an experimental setup we compare the performance of each approach in terms of real-time performance on an embedded smart camera.

The remaining sequences from the PETS2009 dataset that were not used during training and have available ground truth data are used to generate a separate testing dataset. The sequences differ in the time of day, illumination conditions, density and amount of targets, paths of the targets as well as the identity, scale, and appearance of the targets. As such, the performance of the proposed approach is measured objectively. In addition, additional datasets were collected using two different experimental setups to further test whether the approach generalizes to completely different environments.
The proposed approach is compared against three other baselines. All have a detector and tracking component. Comparisons are shown with since single-shot detectors the smaller YOLO\cite{YOLOv2} variant referred to as \textit{tinyYOLO}. In addition, for fairness a CNN (referred to as \textit{PDN}) for pedestrian detection is trained with the YOLO framework on the same dataset as \netnamespace and is developed from scratch for detecting pedestrians on $320\times240$ images. It has a slightly larger feature extraction subnetwork and the latter stages are different to regress the bounding boxes compared to \netname. Comparisons are also made against the widely used SVM-HOG pedestrian \cite{Dalal:Triggs:HOG:2005} detector implemented in OpenCV \cite{opencv_library} still used in many applications (e.g., \cite{Campmany:AutoDriving:HOG:2016,RainDropHOG}). On top of the detectors tracking is applied with Kalman filter which is a common approach used in active tracking \cite{Haj:Beyond_Static:2011,Haj:reacttivePTZ:2010,BewleySimpleTrack}. The incorporated filtering handles bounding box associations, maintains trajectories, and handles the creation and termination of tracks. The kalman filter is prefered since it was not only use in prior works for active tracking but also is less computationally and memory demanding while providing comparable performance \cite{BewleySimpleTrack}. Also note that there is no explicit detection step implemented within the proposed network. As such, it is not possible to directly compare detection performance with standard object detection approaches and metrics. Furthermore, our goal is not detection itself but rather the monitoring performance.
It is important to note that we restrict the comparison to methods followed by previous works and that are more suitable for smart camera applications where the compute-budget is limited. It is worth noting that recent deep-learning-based methods for tracking running on GPU platforms achieve $2-5$ FPS \cite{REVAMP2019}. Hence, methods such as visual trackers that rely on cascade CNNs \cite{SiamCascadeVisTrack} that increase the processing time on top of the detection part are not considered. Another important differentiation to recent works that utilize CNNs for tracking is that in their majority they rely on static camera images and specifically on tracking single objects in the frame. Hence, they are not directly comparable.


\subsection{Performance Metrics}\label{sec:perf_metrics}
The image frames are taken from the PETS2009 dataset however, performance is not measured in the same way since the objective is not to compare with traditional tracking systems that operate on static sensors. Specifically, besides calculating the error between ground truth and predicted motion vector from a set of test images, three metrics are also examined that can holistically evaluate how well an algorithm manages to monitor targets within an area. For all metrics for a target to be considered visible over $50\%$ of the target's body should appear in the frame

1) \textbf{Average Number of Targets in FoV:} The mean number of targets that are monitored over the duration of the sequence. It is the average across the number of visible targets at each frame. It essentially measures how well an algorithm can handle multiple targets within a frame.

2) \textbf{Average Monitoring Time:} The percentage of time for which one or more targets are visible within the FoV of the camera. It essentially measures how well an algorithm does in keeping up with the target(s). This is important aspect as it provides an indication of how likely it is to lose a target and how likely it is to follow a new one once it has entered the FoV.

3)\textbf{Average Distance from Target Centroid:} This is the Euclidean distance of the image center (i.e., camera center) to the centroid of the visible targets (i.e., center of mass). It shows how well an algorithm can keep the targets at the center of its FoV.

\subsection{Evaluation of Learning}\label{saliency}
\begin{figure}[t]
	\centering
	\includegraphics[width=0.99\linewidth]{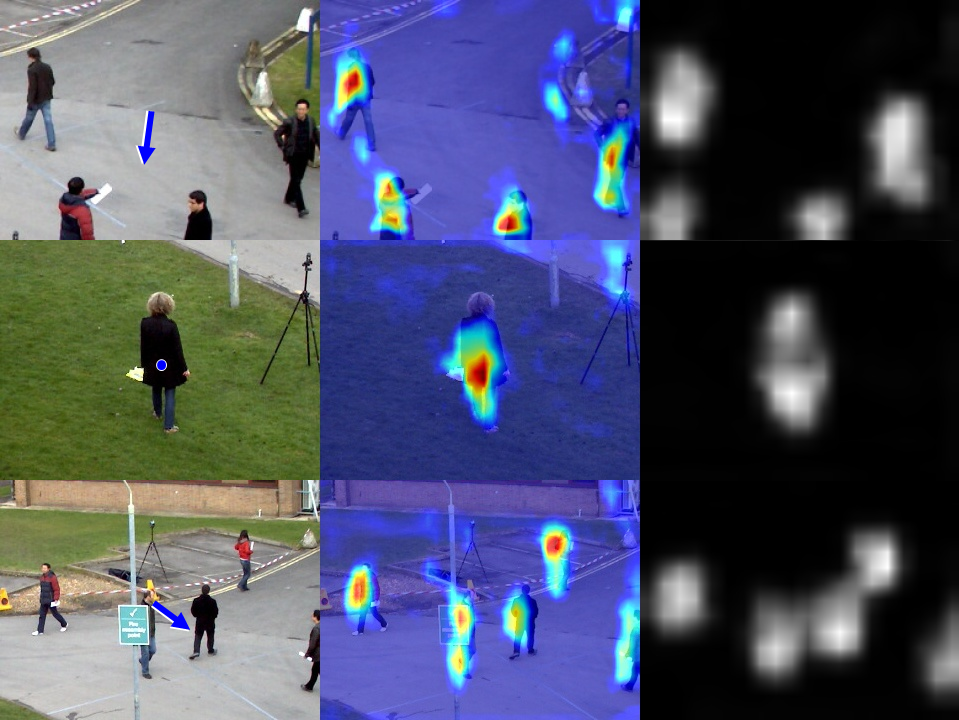}
	\caption{Visualizing what the network learns to look for in images (left) Input Image and direction of motion estimated by \netname. (Center) Visualizations created using Grad-CAM for finding the pixels that mostly influnce the network prediction. (Right) Activation map produced by the controller sub-network by aggregating the features from B5 and B6.(Best viewed in color)}
	\label{fig:deepviz}
\end{figure}

\begin{figure}[t]
	\centering
	\includegraphics[width=0.99\linewidth]{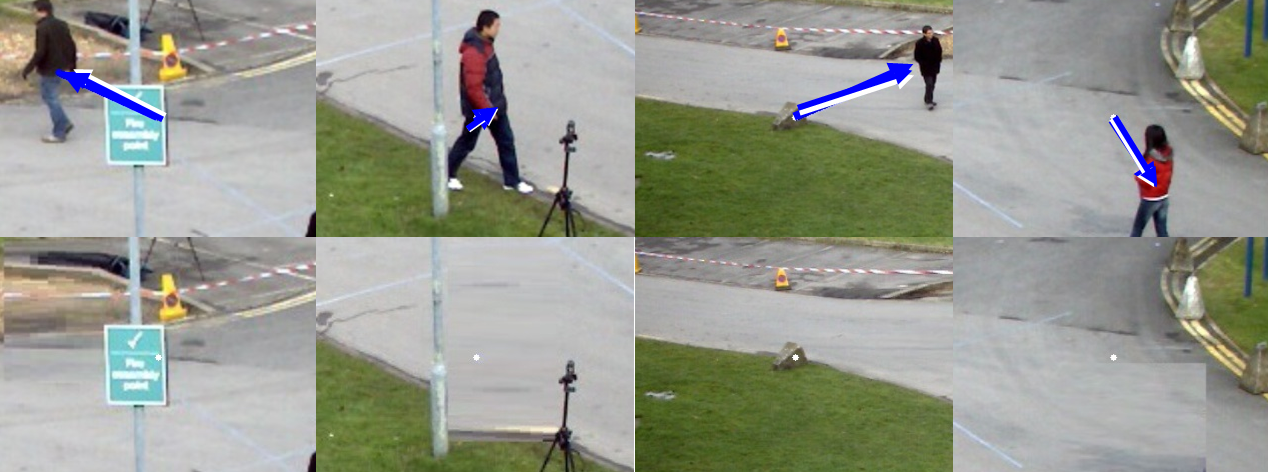}
	\caption{Results on the same image with and without the presence of targets. In the case that no target is present in the image the camera does not move. (Best Viewed in Color)}
	\label{fig:ablation}
\end{figure}

First, a closer look is taken into what features and image regions influence the prediction of \netnamespace and what it has learned to respond to in the various cases. To this end, the Gradient-weighted Class Activation Mapping (Grad-CAM) approach proposed in \cite{Selvaraju.CAM.2017} is employed to generate heat maps of the image regions that mostly influence the control decision by the network. The visualization is with regards to the target outputs of the network. In addition, the activation maps of the feature-extracting sub-network are visualized to understand what features the CNN is looking at (specifically the sum of the spatial averages from blocks B5 and B6). As shown in Fig. \ref{fig:deepviz}, the Grad-CAM visualization the network bases its control output on the targets in the image even if they are partially occluded. Note, that the network is never given the bounding boxes and never explicitly trained to detect people or any other body part but as shown in the activation visualization these are exactly the features that it detects in order to control the camera. Another way to gain confidence in what the network has learned to look for is through an ablation study where image parts are systematically removed to see how relevant they are to the network output. As shown in Fig. \ref{fig:ablation} for images not used during training, the network will only steer the camera to move only when a target is present in the image view. Also notice that it responds even in cases where the target is at different resolutions and not fully visible.

\subsection{Performance in Simulated Environment}

\begin{figure*}[t]
      \centering
      \includegraphics[width=.99\linewidth]{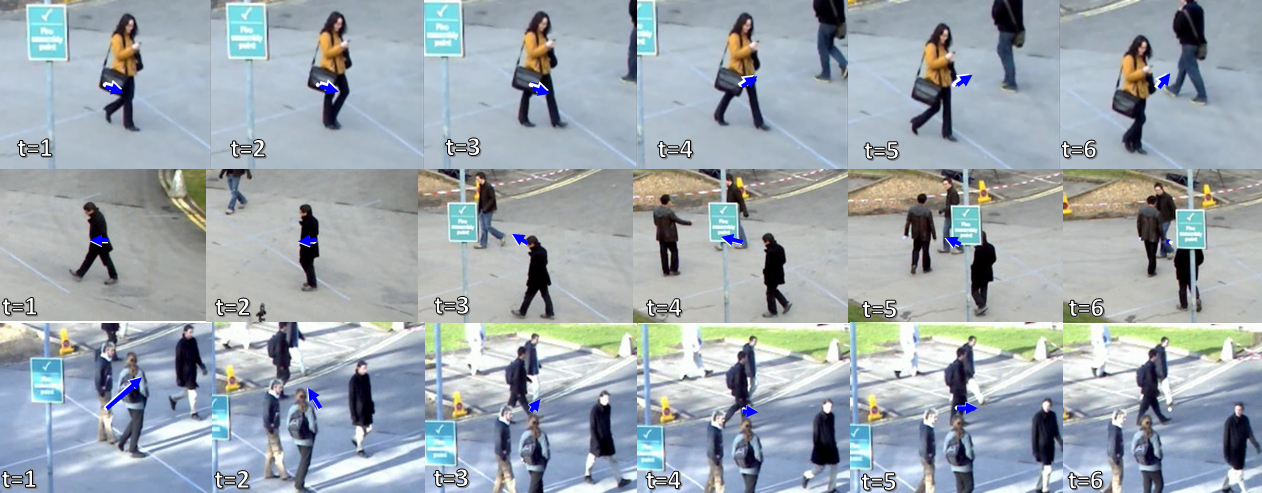}
      \caption{Samples of Sequences: (a) Seq1: Following 1 and then 2 targets. Notice that the network decides to move towards the center of mass of the two targets. (b) Seq2: The network follows a single target then it moves to the center of the group. Also the network does not move away from the partially occluded target. (c) Seq3: Multiple targets present the network adjusts to the movement of incoming targets.}
      \label{fig:seqs}
\end{figure*}

The next experiment involved evaluating all the aforementioned methods in the simulation environment and the performance metrics from Sec. \ref{sec:perf_metrics}. But first it is important to understand why common tracking benchmarks are not well suited for this purpose. The main objective during evaluation is to examine the behaviour of each algorithm and the influence of the control actions on the overall performance. Hence, a camera view needs to be placed within a larger frame in a suitable simulation environment that provides images of moving targets that the virtual camera can follow. In addition, the set of images needs to come with bounding box annotations to calculate whether a target is within the FOV and subsequently measure the different performance metrics described in section \ref{sec:perf_metrics}. Hence to conduct these experiments, $3$ sequences from the PETS2009\cite{PETS2009} dataset were used for which the ground truth is available but have not been used in the training and validation phases. The developed simulation framework outlined in section \ref{sec:simframe} models the camera motion based on the visual input and evaluates its performance. The output of each vision pipeline is a motion vector that will be passed to the simulator to perform the action. These sequences provide an additional challenge as they feature different visual conditions with higher crowd densities and different motion patterns. All approaches start at the same point in every video sequence. The images in the sequences are of $768\times576$ resolution and the virtual camera FoV is set to $320\times240$ for all methods so that there is margin for the camera to move and follow the targets. The first sequence (Seq. 1) has $241$ images and features sparse groups of people walking together, the second (Seq. 2) has $795$ frames and shows individuals independently moving in the area, and the third (Seq. 3) contains $436$ frames of sparse crowd moving randomly in different directions. The camera FoV for all methods is set at the same initial position. In all cases the objective of the camera is to move in such a way as to keep the most number of targets in its FoV. Fig. \ref{fig:seqs} shows some examples from each sequence and the direction of movement chosen by the network. 

\begin{table}[t]
    \begin{center}
    \caption{Error analysis on estimated values.}
    \begin{tabular}{c|c|c|c|c}
    
       \textbf{Method} & \multicolumn{2}{c|}{\textbf{Avg. Error}} & \multicolumn{2}{c}{\textbf{Max Error}} \\
       & $\theta_x$ & $\theta_y$ & $\theta_x$ & $\theta_y$   \\
       \hline\hline
       \netnamespace (Proposed) & \textbf{0.074} & \textbf{0.020} & \textbf{0.315} & \textbf{0.344} \\
       \hline
       SVM-HOG w tracking & 0.170 & 0.086 & 0.917 & 0.784\\
       \hline
       tinyYOLO w tracking & 0.136 & 0.077 & 0.899 & 1.076\\
       \hline
       PDN w tracking & 0.076 & 0.034 & 0.459 & 0.351 \\
    \end{tabular}
    \label{tab:errors}
\end{center}

\end{table}

\begin{table*}[t]
    \begin{center}
    \caption{Monitoring results on 3 different sequences.}
    \begin{tabular}{c|c|c|c|c|c|c|c|c|c}
       
       \textbf{Method} & \multicolumn{3}{c|}{\textbf{Avg. Number of Targets in FoV}} & \multicolumn{3}{c|}{\textbf{Average Monitoring Time}} & \multicolumn{3}{c}{\textbf{Avg. of Target Centroid}} 
       \\
       & Seq. 1 & Seq. 2 & Seq. 3 & Seq. 1 & Seq. 2 & Seq. 3 & Seq. 1 & Seq. 2 & Seq. 3  \\
       \hline\hline
       \netnamespace (Proposed) & \textbf{8} & \textbf{3.3} & \textbf{13} & \textbf{100\%} & \textbf{100\%} & \textbf{100\%} & \textbf{21} & \textbf{18} & \textbf{24} \\
       \hline
       SVM-HOG w tracking & 4.4 & 2.9 & 8 & 77\% & 97\% & 96\% & 41.2 & 33 & 43.9 \\
       \hline
       tinyYOLO w tracking & 5.8 & 3 & 10 & 90\% & 100\% & 100\% & 26.5 & 26.9 & 39.06 \\
       \hline
       PDN w tracking & 5 & 3 & 10.5 & 98\% & 74\% & 100\% & 25.8 & 24.2 & 28.4\\
      
    \end{tabular}
    \label{tab:seq_results}
\end{center}

\end{table*}

\textbf{Performance on Still Images: }First to understand precisely how well each algorithm performs, comparisons are made against the ground truth motion on a per image basis and the overall error for each motion direction is calculated. Table \ref{tab:errors} shows that the proposed CNN achieves the lowest average error as well as the lower maximum error in terms of estimating the camera motion controls. Since still images are used the detection approaches are mostly influenced by the predicted bounding boxes, their size, and the smoothing of the response maps through the non-maximum suppression process. In the case of \netnamespace the only difference is that the moving average is not employed. Overall, as shown in Table \ref{tab:errors}, the proposed approach has a much lower error rate across the two control signals while in the worst case it still has lower error margins.

\textbf{Performance on Continuous Frames: } Results using the metrics defined in \ref{sec:perf_metrics} on continuous frames from the $3$ aforementioned sequences are shown in Table \ref{tab:seq_results} which will also incorporate the temporal aspect and demonstrate the effect of how compounded errors can affect the overall performance.  
First, the average number targets in the FoV is recorded which indicates how well each approach manages to follow a group of targets. The proposed approach manages to outperform the other approaches by following between 1-5 more targets on average as it is not affected by missed bounding box detections. In terms of average monitoring time, \netnamespace is active for the whole duration of the sequences whereas the other approaches lose the targets in some cases. Finally, the average distance of the target center of mass from the FOV center is reported. By keeping targets closer to the center it is less probable that they will exit the FoV. \netnamespace on average keeps targets close to the center with the other approaches having a slightly worse performance in this respect due to some bad localization of bounding boxes. In some cases some false positive detections also contributed to the decreased performance as they introduced jitter to the camera motion. Overall, \netnamespace does not face such challenges as it is not affected by issues related to bad localization of bounding box, errors of the tracker due to retaining outdated bounding box information, multiple overlapping bounding boxes that can cause the camera to remain in an area with less targets, and false positives. The performance gains between \netnamespace and other methods can be attributed end-to-end nature of the approach that associates lower-level-features with control actions. The continuous frames experiments also allowed to understand how the network behaves under different motion patterns and situations. As shown, in Fig. \ref{fig:seqs}-a when a new target enters the FoV the solution will focus on the center of all visible targets until one of them moves away in which case it will focus on the most prominent target(s) in its view. 

\subsection{Use-Case Evaluation}\label{Other}
To evaluate the performance of the proposed approach in the real-world under different circumstances an embedded platform equipped to a smart camera and a UAV to provide on-board processing. In the experimental setup we use the network output that is related to the camera FoV angles and thus directly controls the camera motion. The same exact values to move the camera in the simulated environment are used to move the camera in the real world. It is worth noting that none of data captured from the experimental setups were used during the development and training of the network. As such, these experiments also test the generalization capabilities of the network on novel images taken from completely different sources. The real-world experiments also allow for evaluating the processing and timing aspects which was not possible in the simulated environment.

\subsubsection{Smart Camera Experiment}\label{Embedded}
\begin{figure}[t]
	\centering
	\includegraphics[width=0.99\linewidth]{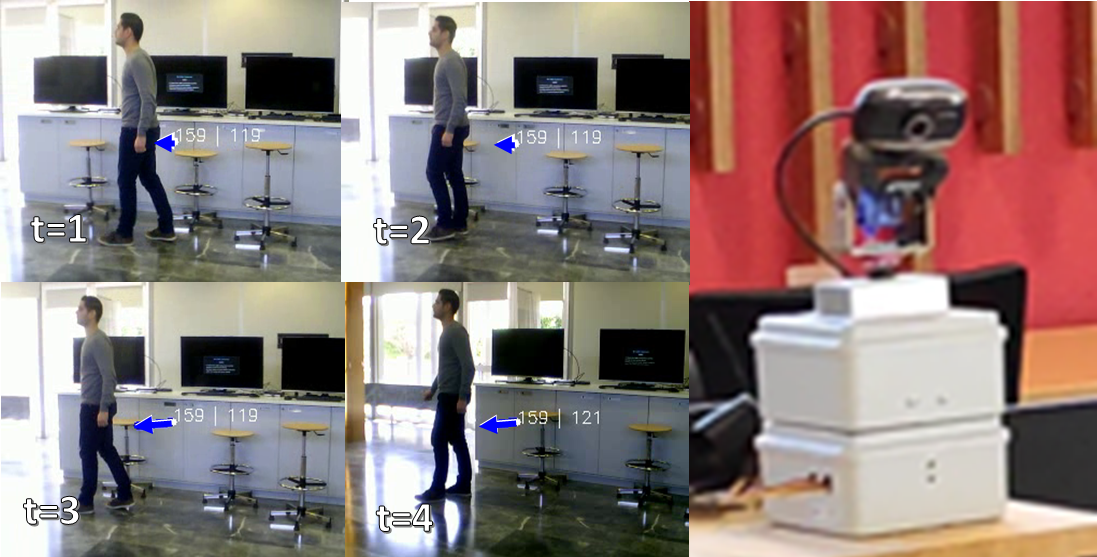}
	\caption{Images from the experimental setup evaluation as well as the embedded smart camera (Best viewed in color)}
	\label{fig:lab_exp}
\end{figure}

\netnamespace is first evaluated on an embedded smart camera based on the Raspberry Pi computer and a webcam \cite{kyrkou:SCN:TCSVT:2018}. The webcam is mounted on a motorized two degrees-of-freedom (DoF) pan-tilt stage. The two angular positions are controlled independently using a corresponding servo motor controlled by the Raspberry Pi. Using off-the shelf motor components the time delay to send the camera control and perform the action was very small since the platform interfaces directly with the motor electronics, thus for the purposes of our experiments it is considered negligible \cite{kyrkou:SCN:TCSVT:2018}. The camera is positioned at a certain height and is tilted with pan range $[-80,\dots,80]$ and tilt range $[0,\dots,45]$. The network sends the motion vector to the camera controller that converts it to angles using Eq. \ref{eq:Motion_Est} and moves the camera head. The indoor environment poses another challenge since there are more reflections, and various background objects in the area. Some results from the lab experiments, as well as the smart camera are shown in Fig. \ref{fig:lab_exp}. Even in environments which are significantly different form the training set, \netnamespace manages to follow the target(s).  

\begin{figure}[t]
	\centering
	\includegraphics[width=0.99\linewidth]{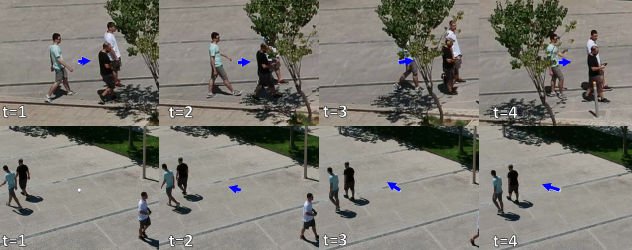}
	\caption{Images taken from the UAV experiment and not used in training. (a)Seq1: The camera using the proposed approach manages to follow the targets even when occluded. (b)Seq2: For a group of targets moving in opposite directions the camera stays with the majority of targets. (Best viewed in color)}
	\label{fig:lab_uav_exp}
\end{figure}

\subsubsection{UAV experiment}\label{Embedded}
To test the algorithm under different conditions we also perform experiments by controlling the camera of a UAV. We assume the same processing platform as with the previous use-case. A series of controlled experiments where carried out with $5$ different targets forming a dataset suitable for UAV target detection and following. The purpose of the experiments was to observe the behaviour of the algorithm as targets perform different motion patterns, such as moving in opposite directions or move behind an obstacle. Examples from these experiments are shown in Fig. \ref{fig:lab_uav_exp}. In the first case, the targets move in a group behind a tree. The algorithm is still able to follow the targets even though they are partially occluded. In the second case, the algorithm is centered at the field of view of the targets and as they diverge it chooses to stay with the majority of them since more features are present in that area of the image. Notice that the targets appear in the two sequences at different resolutions but the algorithm is still able to follow them. This also indicates that the algorithm is rather robust to variations in the target-camera distance. In our experiments good performance was observed when the target(s) were in the range of $10-50m$, i.e., not too close and not too far for the camera. This suites a variety of smart camera applications.

\subsubsection{Benchmarking the computing/processing time}
To benchmark the processing time we measure single frame performance which is more representative for real-time camera applications and calculate the resulting FPS. Hence, batching techniques are not applicable. The same processing platform is considered for both use-cases. As shown in Table \ref{tab:perf}, the proposed approach running on the smart camera platform can achieve a performance of over $10$ frames-per-second (FPS) which make it suitable for surveillance applications using embedded smart cameras \cite{Kulathumani2011:EmbeddedCameras}. The other approaches require considerably more processing time on such low-power platforms ($4$ FPS at best) and hence are not suited for embedded active smart camera applications. The increased latency also hinders the performance of other approaches since targets may drastically change their direction of movement within the time necessary to calculate a new control output. Further \netnamespace requires $\sim40\times$ less memory ($\sim4MB$) compared to tinyYOLO ($\sim177MB$). The nets are also benchmarked on the Titan-XP GPU platform for reference. Notice, that \netnamespace is an order of magnitude faster than the other approaches. 

\begin{table}[h!]
  \begin{center}
    \caption{Computing results based on single frame performance}
    \begin{tabular}{l|c|c|c|c}
       & \netname & PDN & tinyYOLO & SVM-HOG$^1$\\
      \hline
      \hline
      FPS(embedded) & 13 & 4 & 3 & 4\\
      \hline
      FPS(TitanXP) & 367 & 85 & 52 & N/A\\
      \hline
      Memory (MB) & 4 & 18 & 177 & N/A\\
    \end{tabular}
    \label{tab:perf}
  \end{center}
  \text{$^1$No GPU implementation}
\end{table}
\vspace{-3em}
\section{Discussion}\label{discussion}
Overall, the results are encouraging in that \netnamespace learns the camera control task and generalizes to unseen environments. It first builds an internal representation of the object(s) of interest and learns how to move based on the aggregated feature maps. As such, it has learned to detect stationary targets within a single frame and is guided by their positioning and center of mass.

The proposed end-to-end approach is more robust in finding the control actions as it eliminates the need to rely on bounding box predictions which necessitate complex networks, can be noisy and face difficulties with dense targets as illustrated by the results. Such an approach is well suited for scenarios where no one target is more important that the others but persistent monitoring is necessary while following the maximum number of targets within the FoV. As the objective function is implicitly defined through the training set the approach is flexible and can be adapted to different purposes. For example, the action can be weighted by the distance of the targets from the image center as to give more emphasis on targets at the center of the image that have a higher probability of remaining within the image in future time steps. 

Another aspect that is important in practical applications is the response time of the system. The slower it is, the higher the probability that a target will exit its FoV if they suddenly change their motion patterns. The proposed system can tolerate such changes in the order of $0.1$ seconds, and can be further improved. 

The proposed approach can operate alongside other vision subsystems, can provide robust camera control while other subsystems can perform counting, detection and reidentification at different time intervals. \netnamespace can be used to follow a group of people while blobs from the activity map, which is is computed at no additional cost, can provide regions of interest.

\section{Conclusion}

This work investigated how to design a more efficient smart camera controller for active vision applications via end-to-end learning using deep convolutional neural networks. A small CNN architecture referred to as \netnamespace is proposed that maps input images to pan and tilt motion commands. This alleviates the need to have multiple sub-components leading to improved processing times. Even with single image information the proposed network outperforms other multi-stage approaches in terms of monitoring efficiency. Finally, it provides higher frame-rates (a speedup of $\sim4\times$) while being lightweight validating the assumption that end-to-end approaches can lead to smaller and simpler systems. Our results indicate that when appropriately formulated, end-to-end approaches can lead to more efficient systems thus providing a promising avenue for future research. As future work more elaborate architectures will be explored to take advantage of temporal information. In addition, even though this paper has primarily tackled pan and tilt camera motion, the present study can provide a foundation for similarly controlling other parameters such as zoom. 

\section*{Acknowledgment}
The author would like to acknowledge the support of NVIDIA Corporation with the donation of the Titan Xp GPU used for this research.

\bibliographystyle{IEEEtran}
\bibliography{arxiv}

\end{document}